\documentclass{article}

\usepackage{PRIMEarxiv}

\usepackage[utf8]{inputenc} 
\usepackage[T1]{fontenc}    
\usepackage{hyperref}       
\usepackage{url}            
\usepackage{booktabs}       
\usepackage{amsfonts}       
\usepackage{nicefrac}       
\usepackage{microtype}      
\usepackage{fancyhdr}       
\usepackage{graphicx}       
\graphicspath{{media/}}     

\usepackage{subcaption}
\usepackage{threeparttable}
\usepackage[table]{xcolor}
\usepackage{tabularx}
\usepackage{array}
\usepackage{float}
\usepackage{algorithm}
\usepackage{algorithmic}
\usepackage{amsmath}
\usepackage{amssymb}
\usepackage{mathtools}
\usepackage{amsthm}
\usepackage{iftex}
\usepackage[capitalize,noabbrev]{cleveref}
\usepackage[disable,textsize=tiny]{todonotes}

\ifPDFTeX
\DeclareUnicodeCharacter{2011}{\nobreakdash-}
\DeclareUnicodeCharacter{2013}{--}
\DeclareUnicodeCharacter{2014}{---}
\fi

\theoremstyle{plain}

\theoremstyle{definition}

\theoremstyle{remark}

\title{ChemHyperMag: Physics-informed magnetic hypergraph learning improves molecular ADMET prediction}

\author{
\normalfont
Hexiao Ding\textsuperscript{1,*},
Hongzhao Chen\textsuperscript{1,*},
Jing Lan\textsuperscript{1,*},
Yufeng Jiang\textsuperscript{1},
Zihong Luo\textsuperscript{2},
Zehua Xiong\textsuperscript{3},
Tianlong Ruan\textsuperscript{4},\\
Yunlin Mao\textsuperscript{1},
Nga Chun Ng\textsuperscript{1},
Gwing Kei Yip\textsuperscript{1},
Gerald W.Y. Cheng\textsuperscript{1},
Kate Inyoung Oh\textsuperscript{1},
Jing Cai\textsuperscript{1},\\
Liang-Ting Lin\textsuperscript{1},
Jung Sun Yoo\textsuperscript{1,\#}\\[0.5em]
\textsuperscript{1}Department of Health Technology and Informatics, Hong Kong Polytechnic University, Hong Kong SAR, China\\
\textsuperscript{2}University of Liverpool, Liverpool, United Kingdom\\
\textsuperscript{3}Shanghai Jiao Tong University, Shanghai, China\\
\textsuperscript{4}Fudan University, Shanghai, China\\[0.5em]
\textsuperscript{*}Equal contribution. \textsuperscript{\#}Corresponding author.\\
\texttt{jungsun.yoo@polyu.edu.hk}
}

\begin{document}
\maketitle

\begin{abstract}
Accurate prediction of ADMET (Absorption, Distribution, Metabolism, Excretion, and Toxicity) is important for drug discovery. Most predictors use undirected molecular graphs and pairwise edges. This choice misses asymmetric interactions, nonreversible dynamics, and motif level effects from functional groups and ring systems. We propose ChemHyperMag for multitask ADMET prediction under missing labels. ChemHyperMag builds a functional group hypergraph from rings, BRICS fragments, Bemis-Murcko scaffolds, and bonds. It also defines a potential driven nonreversible flow guided by electronegativity and Gasteiger partial charges. The resulting circulation is encoded by a Hermitian magnetic Laplacian and processed with a magnetic Chebyshev encoder. We perturb magnetic phases to form stochastic views and train with an InfoNCE objective. Experiments on multiple ADMET benchmarks show improvements over recent methods with fewer labeled samples and no conformers. ChemHyperMag is scalable and provides interpretable directional signals through its magnetic phases. This work was accepted by Proceedings of the AI4Physics Workshop at the 43 rd International Conference on Machine Learning (AI4Physics@ICML 2026, \url{https://openreview.net/forum?id=KI9UxQtJlM}). The code can be found at \url{https://github.com/1anj/ChemHyperMag}.
\end{abstract}


\section{Introduction}

The prediction of ADMET (Absorption, Distribution, Metabolism, Excretion, and Toxicity) is a key question in the discovery of small-molecule drugs. It determines which candidates justify costly synthesis and assays. ADMET benchmarks in machine learning encompass multi‑task learning, heterogeneous label semantics, and pervasive missingness \cite{wu2018moleculenet, Huang2022artificial}. Currently, many workflows begin with SMILES \cite{weininger1988smiles}, which are modeled as sequences or transformed into two‑dimensional atom-bond graphs by RDKit \cite{bento2020open}. Molecular encoders underperform in this regime because their representations lack comprehensive structural detail and fail to capture the physical and chemical nature of molecules.

Sequence encoders built on SMILES provide only indirect access to molecular structure, so functional regularities must be inferred from sparse and heterogeneous endpoints \cite{weininger1988smiles}. Message passing on bond graphs exposes topology, yet it is typically defined through undirected pairwise interactions. This formulation ensures invariance to edge reversal but neglects bond polarity and donor–acceptor roles, thus higher-order motifs are merely approximated via local aggregation \cite{yang2019analyzing}. However, the abstraction of these representations is misaligned with the physical drivers behind many ADMET endpoints. Transport and partitioning are shaped by local polarity fields induced by electronegativity imbalance and charge separation \cite{pauling1960nature,gasteiger1980iterative}, while conjugated substructures exhibit directional redistribution and cyclic electronic circulation that cannot be expressed by undirected diffusion. Geometry-based and multimodal models can recover part of this signal. However, they rely on conformer generation and large-scale three-dimensional supervision, which raises data and compute demands and often conflicts with high-throughput ADMET settings \cite{zhou2023unimol, lu2024data,ji2024exploring}. 
The challenge is to design a SMILES-based pipeline that captures higher-order motif structures. At the same time, it should embed physically and chemically grounded directionality to enhance learning processes.

\begin{figure}[!t]
    \centering
    \includegraphics[width=1.0\linewidth]{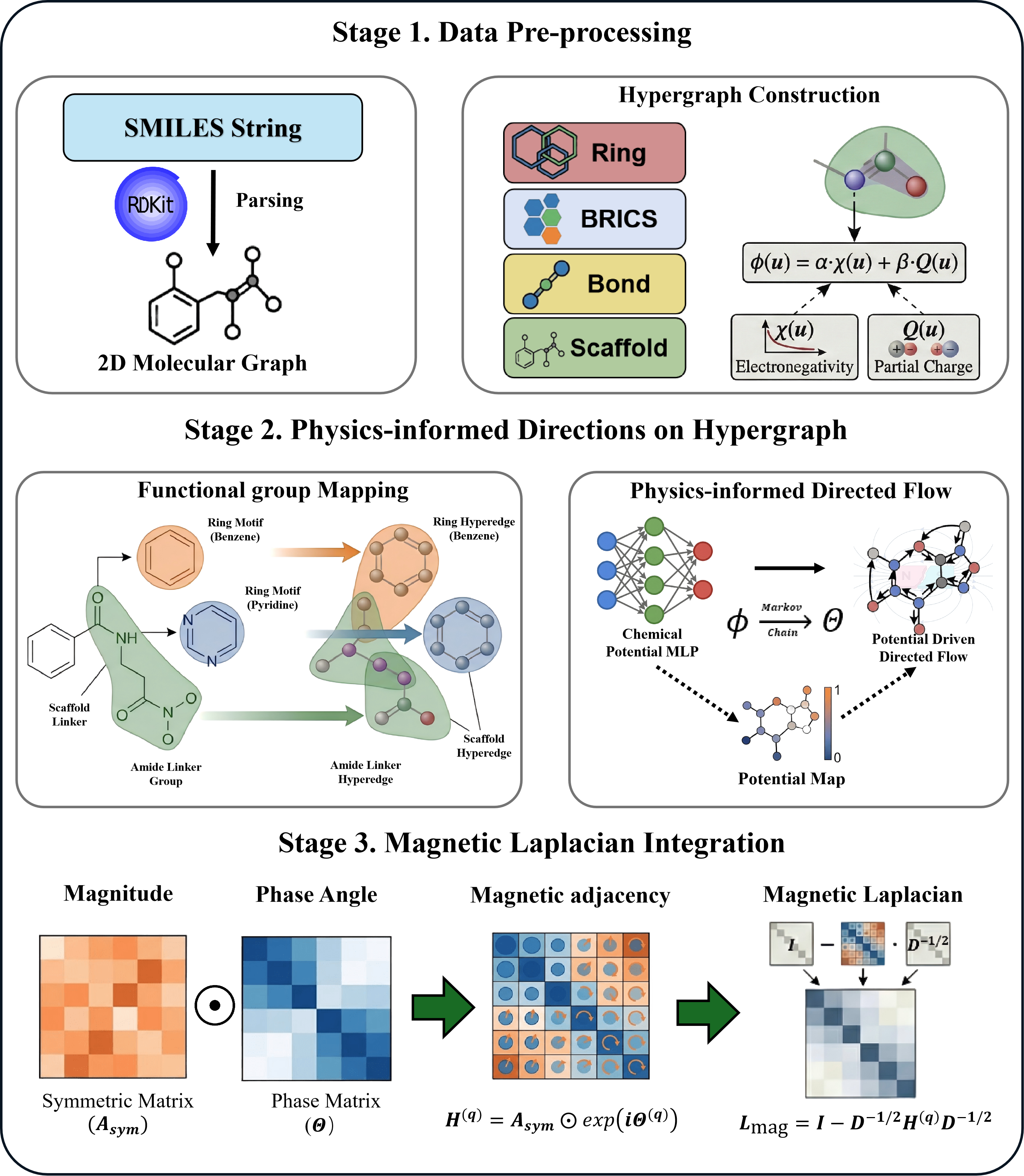}
    \caption{Workflow of ChemHyperMag. \textbf{\textit{Stage 1}} parses SMILES into a 2D molecular graph with RDKit and computes atom electronegativity and Gasteiger partial charges together with structural annotations. \textbf{\textit{Stage 2}} constructs a functional group hypergraph from rings, BRICS fragments, scaffolds, and bonds, then defines a chemical potential that induces directed flow under the hypergraph constraints. \textbf{\textit{Stage 3}} encodes this directionality with a magnetic phase, forms the magnetic adjacency, and builds the normalized magnetic Laplacian for spectral propagation.}
    \label{fig:1}
\end{figure}

We propose ChemHyperMag to address this issue. It introduces directionality by integrating physically and chemically grounded energy-structure coupling. We start from the standard 2D molecular graph extracted from SMILES, which fixes the atom set and covalent connectivity. We then lift this graph to a functional group hypergraph that augments pairwise bonds with motif-defined higher-order neighborhoods, so that propagation is supported not only on covalent edges but on atom pairs that co-occur within chemically meaningful motifs. On this hypergraph-supported neighborhood we define a potential-driven, nonreversible diffusion. Each atom is assigned a chemical potential computed from its electronegativity and partial charge, and transitions are biased along potential gradients. The resulting Markov operator induces a stationary flow that decomposes into a symmetric component encoding diffusion strength and an antisymmetric component encoding circulation. We encode this circulation with a magnetic Laplacian, where the complex phase represents cyclic currents while the operator remains Hermitian, enabling stable learning while retaining directional information \cite{fanuel2017magnetic,zhang2021magnet,benko2025hypermagnet}. To improve robustness under missing labels, we construct stochastic views by perturbing the magnetic phase while keeping the hypergraph fixed. We then optimize an InfoNCE objective to align these views at the operator level, following directed contrastive learning \cite{tong2021directed}. This design removes the need for conformers, but it introduces sparse hypergraph construction and complex-valued propagation. \textit{\textbf{Figure~\ref{fig:1}}} illustrates the concept and workflow of ChemHyperMag.

The main contributions of this work are summarized as follows. First, we build a SMILES-derived functional group hypergraph that encodes motif-level neighborhoods beyond pairwise bonds. Second, we introduce a physicochemically-grounded nonreversible Markov flow on this hypergraph and encode its directional bias and circulation with a Hermitian magnetic Laplacian through complex phases. Third, we propose magnetic phase perturbation as operator-level view generation for InfoNCE regularization in multitask ADMET with missing labels.

\section{Related Work}
\label{sec_related}

ADMET prediction is framed as multi-endpoint learning with heterogeneous labels and strong emphasis on standardized splits and metrics, as in MoleculeNet \cite{wu2018moleculenet} and the Therapeutics Data Commons \cite{Huang2022artificial}.

\subsection{Statistical Learning Methods}
Quantitative Structure-Activity Relationship (QSAR) baselines remain central for ADMET. Extended-connectivity fingerprints used physicochemical descriptors and fingerprints with linear models or tree ensembles \cite{rogers2010extended}. The other QSAR models such as pkCSM \cite{pires2015pkcsm}, SwissADME \cite{daina2017swissadme}, vNN-ADMET \cite{schyman2017vnn}, and ADMETboost \cite{tian2022admetboost} serve as rule- and descriptor-based screening tools. 

\subsection{Deep Learning Methods}
The next-generation ADMET platforms combine these ingredients with learned-models and curated endpoints at scale. ADMETlab~2.0 \cite{xiong2021admetlab} and 3.0 \cite{fu2024admetlab} and ADMET-AI \cite{swanson2024admet} are pioneer works. Besides, end-to-end representation learning largely follows message passing on atom-bond graphs, with PotentialNet \cite{feinberg2018potentialnet} and the D-MPNN family implemented in Chemprop \cite{yang2019analyzing, heid2023chemprop}. Notably, Chemprop v2 is now serving as a widely adopted ADMET backbone \cite{graff2025chemprop}. Moreover, recent advances in transformer architectures and 3D pretraining frameworks, such as Graphormer \cite{ying2021transformers} and Uni-Mol \cite{zhou2023unimol}, have shown improved capacity to address the long-context problem and capture geometric effects in molecular ADMET prediction. Current deep learning models increasingly leverage endpoint structure and analogue comparisons in ADMET prediction. Notable strategies include adaptive auxiliary task selection in MTGL-ADMET \cite{du2023admet} and pairwise difference learning in DeepDelta \cite{fralish2023deepdelta}, while OmniMol emphasizes multi-task robustness under imperfect labeling conditions \cite{wang2025unified}.

\subsection{Physics-informed Methods}
Physics‑informed ADMET models incorporate physicochemical priors by supervising molecular representations with electronic or thermodynamic features beyond connectivity. Recent work uses quantum chemistry supervision to pretrain graph transformers on atom-in-a-molecule properties and transfers the resulting encoders to ADMET endpoints \cite{kim2024quantum,fallani2025pretraining}. Physics-informed methods improve model performance and sample efficiency when labels are sparse or missing.

\subsection{Hypergraph and Magnetic Laplacian}

\subsubsection{Hypergraph Neural Networks}

Hypergraph neural networks generalize graph message passing by defining transmission over node-hyperedge incidence structures \cite{feng2019hypergraph}. Surveys have organized existing approaches into convolutional, attention, and spectral formulations \cite{yang2025recent}. For undirected hypergraphs, multiset-based encoders such as AllSet learn permutation-invariant node-hyperedge aggregation and unify several incidence variants \cite{chien2022allset}. In parallel, sheaf hypergraph networks enrich hyperedge states and produce more expressive Laplacians for modeling higher-order diffusion \cite{duta2023sheaf}. Directed hypergraphs assign each hyperedge a tail and a head, thereby inducing nonreversible dynamics and asymmetric Laplacians. DHGNN introduces a directed-hypergraph Laplacian for convolution and link prediction \cite{ma2024dhgnn}, while DSHN unifies directed and undirected learning through a complex-valued directed sheaf Laplacian \cite{mule2025directional}. In molecular learning, hyperedges are used to represent rings, fragments, and reaction environments \cite{kajino2019molecular,chen2024molecular}. Functional-group hypergraphs have been shown to improve toxicity prediction and related ADMET endpoints in MolHGCN \cite{chen2021hypergraph}, while molecular hypergraph message passing captures irregular higher-order conjugation beyond conventional bond graphs \cite{chen2024molecular}.

\subsubsection{Magnetic Laplacian}
Magnetic Laplacians encode edge orientation in complex phases while remaining Hermitian, which enables spectral learning under nonreversible dynamics \cite{fanuel2017magnetic}. MagNet introduces a charge-parameterized magnetic Laplacian for directed representation learning \cite{zhang2021magnet}, and MSGNN generalizes it to signed digraphs through a magnetic signed Laplacian with Chebyshev polynomial filters \cite{he2022msgnn}. SD-GCN derives a related Hermitian magnetic Laplacian for signed directed graphs and uses it for spectral convolution \cite{ko2023spectral}. HyperMagNet extends magnetic operators to hypergraphs by defining an edge-dependent random walk on the incidence bipartite graph and constructing a magnetic hypergraph Laplacian with learnable charge \cite{benko2025hypermagnet}.

\subsection{Graph Contrastive Learning (CL)}

\subsubsection{Graph CL without Directions}

Graph contrastive learning (GCL) learns robust graph representations by promoting representation scattering, often via agreement between augmented views or distribution discrimination \cite{he2024exploitation}. Representative frameworks such as GraphCL \cite{you2020graph} have standardized objectives and augmentation pipelines. Follow-up research improves view construction and loss formulation using adversarial and prototypical strategies \cite{suresh2021adversarial, lin2022prototypical}. Empirical analyses also identify scenarios where augmentations enhance representation learning or cause collapse \cite{zhu2021an}. For molecular ADMET predictions, labeled endpoints are heterogeneous and often limited. Recent GCL frameworks enhance data efficiency by tailoring chemically-valid perturbations and enforcing cross-view alignment. MolCLR \cite{wang2022molecular}, GraphMVP \cite{liu2022pretraining}, and 3DGCL \cite{moon20233d} advance 2D-3D contrast; ATMOL \cite{liu2022attention} introduces attention-guided masking; and ACML \cite{wang2025advancing} develops multimodal asymmetric alignment. Collectively, these methods achieve consistent gains on drug-like benchmarks with ADMET endpoints.

\subsubsection{Graph CL with Directions}

Directed Graph CL avoids orientation loss by perturbing graph operators. DiGCL leverages directed Laplacian perturbations with dynamic view sampling \cite{tong2021directed}, and UGCL extends this framework to unify Laplacian perturbations across undirected, directed, and signed graphs \cite{ko2023universal}. For signed directed structure, SDGCL \cite{ko2023signed} introduces Laplacian augmentation in magnetic space and S2-DiGCL further couples path-based spatial views with operator-level perturbations \cite{su2025toward}. However, molecular representations are generally treated as undirected graphs, and directed graph contrastive learning remains unexplored in ADMET prediction.

\section{Preliminaries}

\subsection{Problem Definition}
\label{subsec_problem}
Let $\mathcal{G}$ be the set of molecular graphs and let $T\in\mathbb{N}$ be the number of prediction tasks. The training set is
\begin{equation}
\mathcal{D}
=
\bigl\{(G_i,y_i,m_i)\bigr\}_{i=1}^N
\subseteq
\mathcal{G}\times\mathbb{R}^T\times\{0,1\}^T,
\label{eq_dataset}
\end{equation}
where $G_i\in\mathcal{G}$ is the $i$th molecule, $y_i=(y_{i1},\dots,y_{iT})\in\mathbb{R}^T$ is its multitask label vector with $y_{it}$ the ground truth target for task $t$, and
$m_i=(m_{i1},\dots,m_{iT})\in\{0,1\}^T$ is the observation mask with $m_{it}=1$ if $y_{it}$ is observed and $m_{it}=0$ otherwise. A model $f_\theta:\mathcal{G}\to\mathbb{R}^T$ maps an input $G$ to a prediction vector $f_\theta(G)$, whose $t$th component $f_\theta(G)_t$ predicts task $t$.

On a mini batch $\mathcal{B}\subset\mathcal{D}$ the supervised multitask loss is
\begin{equation}
\mathcal{L}_{\mathrm{sup}}
=
\frac{1}{|\mathcal{B}|}
\sum_{(G,y,m)\in\mathcal{B}}
\sum_{t=1}^T
m_t\,\ell_t\!\bigl(f_\theta(G)_t,\,y_t\bigr),
\label{eq_generic_supervised}
\end{equation}
where $\ell_t$ is the loss for task $t$. ChemHyperMag implements $f_\theta$ as a shared magnetic hypergraph encoder with task specific heads and combines \eqref{eq_generic_supervised} with a contrastive term.

\subsection{Functional Group Hypergraphs}
\label{subsec_mol_hyper}

\subsubsection{Molecular Graphs}
A two dimensional molecular graph is $G=(V,B,X)$, where $V=\{1,\dots,n\}$ is the set of atoms, $B\subseteq V\times V$ is the set of undirected covalent bonds, and $X\in\mathbb{R}^{n\times d}$ is the node feature matrix with rows $x_u\in\mathbb{R}^d$ for $u\in V$.

The bond graph $(V,B)$ encodes connectivity and bond distance. For $u,v\in V$, a bond path of length $k$ from $u$ to $v$ is a sequence $(u_0,\dots,u_k)$ with $u_0=u$, $u_k=v$, and $(u_{r-1},u_r)\in B$ for all $r\in\{1,\dots,k\}$. Define the bond distance
\begin{equation}
\begin{aligned}
\operatorname{dist}_B(u,v) &= \min \bigl\{ k\in\mathbb{N}_0 : \exists\,(u_0,\ldots,u_k) \text{ s.t. } \\
&\quad u_0=u,\ u_k=v,\ (u_{r-1},u_r)\in B, \forall r\in[k] \bigr\},
\end{aligned}
\label{eq_bond_distance}
\end{equation}
with $\operatorname{dist}_B(u,v)=+\infty$ if no such path exists. In particular, $u$ and $v$ are connected if and only if $\operatorname{dist}_B(u,v)<+\infty$.

\subsubsection{Hypergraphs}
To represent functional groups and structural motifs, we lift the molecule to a hypergraph $\mathcal{H}=(V,E)$ with the same node set $V$ and hyperedges $E=\{e_1,\dots,e_m\}$, where each $e_j\subseteq V$. The incidence matrix is denoted by $\mathbf{M}\in\{0,1\}^{n\times m}$ with entries
\begin{equation}
\mathbf{M}_{u,j}
=
\begin{cases}
1, & u\in e_j,\\
0, & u\notin e_j,
\end{cases}
\label{eq_incidence}
\end{equation}
for $u\in V$ and $j\in\{1,\dots,m\}$. In ChemHyperMag, each hyperedge $e_j$ corresponds to a ring, a BRICS fragment, a Bemis-Murcko scaffold, or a bond hyperedge.

Throughout, $u,v\in V$ denote atoms, and an ordered pair $(u,v)$ denotes a directed transition from $u$ to $v$. For $e\in E$ and $u\in e$, define
\begin{equation}
N_e(u)=\{v\in e \mid v\neq u\},
\label{eq_within_edge_neighborhood}
\end{equation}
and the hypergraph neighborhood
\begin{equation}
N(u)=\bigcup_{e\in E\,:\,u\in e} N_e(u).
\label{eq_hyper_neighborhood}
\end{equation}
Thus $v\in N(u)$ means that $u$ and $v$ co-occur in at least one hyperedge. All Markov transitions in Section~\ref{subsec_markov_magnetic} are supported on ordered pairs $(u,v)$ with $v\in N(u)$.

Each atom $u\in V$ carries a scalar chemical potential
\begin{equation}
\phi(u)=\alpha\,\chi(u)+\beta\,Q(u),
\label{eq_chem_potential}
\end{equation}
where $\chi(u)$ is the Pauling electronegativity of the atom at $u$ and $Q(u)$ is its Gasteiger partial charge. The scalars $\alpha$ and $\beta$ are learned parameters, and $\phi$ biases the nonreversible Markov dynamics.

\subsection{Magnetic Markov Operators}
\label{subsec_markov_magnetic}

\subsubsection{Ergodic Markov Chains}
Let $V=\{1,\dots,n\}$ be the atom set of Section~\ref{subsec_mol_hyper}, which we also use as the finite state space of the Markov chain. A row stochastic matrix is $P\in[0,1]^{n\times n}$ with nonnegative entries and
\begin{equation}
P\mathbf{1}=\mathbf{1},
\label{eq_row_stochastic}
\end{equation}
where $\mathbf{1}$ is the all ones vector. We interpret $P_{uv}$ as the one step transition probability from state $u$ to state $v$. A distribution $\pi\in\mathbb{R}^n$ with $\pi_u\ge 0$ and $\mathbf{1}^\top\pi=1$ is stationary for $P$ if $\pi^\top P=\pi^\top$.

To guarantee ergodicity, we add a teleportation term. The same teleporting random walk is used in directed graph spectral learning~\cite{tong2021directed} and in magnetic hypergraph operators~\cite{benko2025hypermagnet}. Given a base transition matrix $P$ and a distribution $\nu\in\mathbb{R}^n$ with full support, meaning $\nu_u>0$ for all $u\in V$, for $\eta\in(0,1)$ define
\begin{equation}
P^{(\eta)}=(1-\eta)P+\eta\,\mathbf{1}\,\nu^\top.
\label{eq_teleport}
\end{equation}
Since $(P^{(\eta)})_{uv}\ge \eta\nu_v>0$ for all $u,v\in V$, the chain is irreducible and aperiodic, hence admits a unique stationary distribution. We denote it by $\pi^{(\eta)}$, and when $\eta$ is fixed we write $\pi$.

\subsubsection{Phase Map and Magnetic Adjacency}
Let $A\in\mathbb{R}^{n\times n}$ be a real matrix supported on the allowed ordered pairs, meaning $A_{uv}=0$ whenever $v\notin N(u)$. Define its symmetric and skew symmetric parts
\begin{equation}
A^{\mathrm{sym}}=\frac{A+A^\top}{2},
\qquad
A^{\mathrm{skew}}=\frac{A-A^\top}{2}.
\label{eq_sym_skew}
\end{equation}
For $u,v\in V$ with $A^{\mathrm{sym}}_{uv}>0$ define
\begin{equation}
R_{uv}=\frac{A^{\mathrm{skew}}_{uv}}{A^{\mathrm{sym}}_{uv}+\varepsilon},
\qquad
\Theta^{(q)}_{uv}=2\pi q\,\tanh(R_{uv}),
\label{eq_phase_map}
\end{equation}
with $\varepsilon>0$ and $q\ge 0$. Then $\Theta^{(q)}_{uv}=-\Theta^{(q)}_{vu}$ and $\Theta^{(q)}_{uu}=0$.

Define $\Xi^{(q)}\in\mathbb{C}^{n\times n}$ by
\begin{equation}
\Xi^{(q)}_{uv}
=
\begin{cases}
\exp\!\bigl(\mathrm{i}\,\Theta^{(q)}_{uv}\bigr), & A^{\mathrm{sym}}_{uv}>0,\\
1, & A^{\mathrm{sym}}_{uv}=0,
\end{cases}
\label{eq_phase_factor}
\end{equation}
where the choice on $\{(u,v)\mid A^{\mathrm{sym}}_{uv}=0\}$ is irrelevant since $H^{(q)}_{uv}=0$ there. For $A^{\mathrm{sym}}_{uv}>0$, $|\Xi^{(q)}_{uv}|=1$ and $\Xi^{(q)}_{uv}=\overline{\Xi^{(q)}_{vu}}$, hence $\Xi^{(q)}$ is Hermitian and generally not real.

The magnetic adjacency is
\begin{equation}
H^{(q)}=A^{\mathrm{sym}}\odot \Xi^{(q)},
\qquad
(A\odot B)_{uv}=A_{uv}B_{uv}.
\label{eq_magnetic_adjacency}
\end{equation}

\subsubsection{Magnetic Laplacian}
Let $d_u=\sum_{v\in V}A^{\mathrm{sym}}_{uv}$ and $D=\operatorname{diag}(d_1,\dots,d_n)$. Assume $d_u>0$ for all $u\in V$. The normalized magnetic Laplacian is
\begin{equation}
L_{\mathrm{mag}}=I-D^{-1/2}H^{(q)}D^{-1/2}.
\label{eq_magnetic_laplacian}
\end{equation}

\medskip
\noindent\textbf{Proposition 1.}
$L_{\mathrm{mag}}$ is Hermitian and positive semidefinite.\\
\noindent\emph{Proof.}
Since $A^{\mathrm{sym}}$ is real symmetric and $\Xi^{(q)}$ is Hermitian, the Hadamard product
$H^{(q)}=A^{\mathrm{sym}}\odot\Xi^{(q)}$ satisfies $H^{(q)}_{uv}=\overline{H^{(q)}_{vu}}$, hence $H^{(q)}$ is Hermitian. The matrix $D^{-1/2}$ is real diagonal, hence Hermitian, so $L_{\mathrm{mag}}$ is Hermitian.
For $z\in\mathbb{C}^n$ define $\widehat z_u=z_u/\sqrt{d_u}$. Using $H^{(q)}_{uv}=A^{\mathrm{sym}}_{uv}\Xi^{(q)}_{uv}$ and $|\Xi^{(q)}_{uv}|=1$, one obtains
\begin{equation}
z^\ast L_{\mathrm{mag}} z
=
\frac12
\sum_{u,v\in V}
A^{\mathrm{sym}}_{uv}\,
\bigl|\widehat z_u-\Xi^{(q)}_{uv}\widehat z_v\bigr|^2
\ge 0,
\label{eq_rayleigh_magnetic}
\end{equation}
which proves positive semidefiniteness.

\subsection{Contrastive Objectives}
\label{subsec_contrastive}
An InfoNCE objective is applied to pairs of stochastic views of the same molecule~\cite{oord2018representation}. Let $\mathcal{X}$ be the space of molecular inputs. A stochastic augmentation kernel $\mathcal{T}$ maps $x\in\mathcal{X}$ to a random view $\widetilde x\sim\mathcal{T}(\cdot\mid x)$. An encoder $f_\theta:\mathcal{X}\to\mathbb{R}^p$ produces embeddings $h=f_\theta(\widetilde x)$. For each $x$, sample two views $\widetilde x^{(1)},\widetilde x^{(2)}\sim\mathcal{T}(\cdot\mid x)$ and set $h^{(k)}=f_\theta(\widetilde x^{(k)})$.

For a mini batch of size $B$, write $(h_i^{(1)},h_i^{(2)})$ for the pair from the $i$th molecule. With cosine similarity
\begin{equation}
\operatorname{sim}(u,v)=\frac{\langle u,v\rangle}{\|u\|_2\,\|v\|_2},
\label{eq_cosine_sim}
\end{equation}
and temperature $\tau>0$, define the batchwise scores
\begin{equation}
s_{ij}^{12}=\operatorname{sim}\!\bigl(h_i^{(1)},h_j^{(2)}\bigr),
\qquad
s_{ij}^{21}=\operatorname{sim}\!\bigl(h_i^{(2)},h_j^{(1)}\bigr).
\label{eq_scores}
\end{equation}
The symmetric InfoNCE loss is
\begin{equation}
\mathcal{L}_{\mathrm{NCE}}
=
\frac{1}{2B}
\sum_{i=1}^B
\bigl(\ell_i^{(1)}+\ell_i^{(2)}\bigr),
\label{eq_info_nce}
\end{equation}
where
\begin{align}
\ell_i^{(1)}
&=
-\log
\frac{\exp\!\bigl(s_{ii}^{12}/\tau\bigr)}
{\sum_{j=1}^B \exp\!\bigl(s_{ij}^{12}/\tau\bigr)},
\label{eq_info_term_1}
\\
\ell_i^{(2)}
&=
-\log
\frac{\exp\!\bigl(s_{ii}^{21}/\tau\bigr)}
{\sum_{j=1}^B \exp\!\bigl(s_{ij}^{21}/\tau\bigr)}.
\label{eq_info_term_2}
\end{align}
In ChemHyperMag, $\mathcal{T}$ perturbs $\Xi^{(q)}$ while keeping the molecular hypergraph $\mathcal{H}=(V,E)$ fixed, and $\mathcal{L}_{\mathrm{NCE}}$ is combined with $\mathcal{L}_{\mathrm{sup}}$ in \textit{\textbf{Equation \eqref{eq_generic_supervised}}}.


\section{Methodology}

\begin{figure*}[t]
    \centering
    \includegraphics[width=1.0\textwidth]{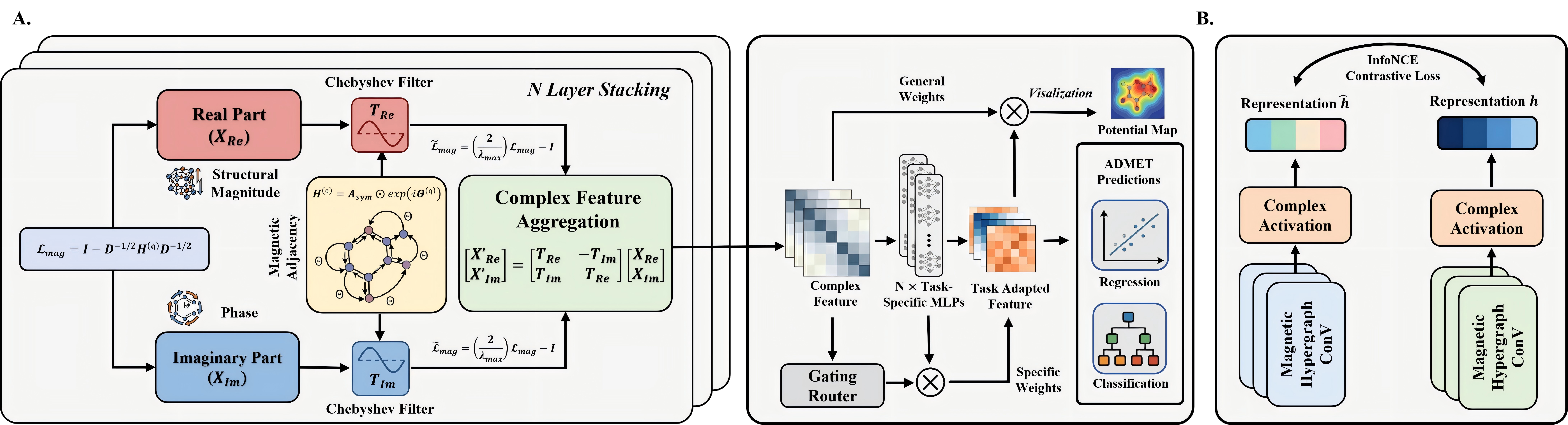}
    \caption{The model framework of ChemHyperMag. \textbf{A.} Internal structure of the Magnetic Chebyshev Encoder and subsequent gating mechanism; \textbf{B.} Brief overview of contrastive learning.}
    \label{fig:2}
\end{figure*}

ChemHyperMag combines a functional group hypergraph representation with magnetic spectral propagation for multitask molecular prediction. Hypergraph construction and notation follow Section~\ref{subsec_mol_hyper}. The encoder instantiates the magnetic operator of Section~\ref{subsec_markov_magnetic} from atom level chemical potential in \textit{\textbf{Equation \eqref{eq_chem_potential}}}, applies a Chebyshev spectral filter on the resulting complex magnetic Laplacian in \textit{\textbf{Equation \eqref{eq_magnetic_laplacian}}}, and produces real node embeddings used by task specific pooling and predictors. Training minimizes the masked supervised loss in \textit{\textbf{Equation \eqref{eq_generic_supervised}}} and the contrastive objective in \textit{\textbf{Equation \eqref{eq_info_nce}}}. The model framework of ChemHyperMag is shown in \textit{\textbf{Figure~\ref{fig:2}}}.

\subsection{Functional Group Hypergraph}

For each molecule $G=(V,B,X)$, we construct the hypergraph $\mathcal{H}=(V,E)$ as in Section~\ref{subsec_mol_hyper}, with hyperedges from SSSR rings~\cite{figueras1996ring}, BRICS fragments~\cite{degen2008art}, the Bemis-Murcko scaffold~\cite{bemis1996properties}, and bond hyperedges $\{u,v\}$ for all $(u,v)\in B$. We implement $\mathcal{H}$ as an incidence graph whose node set augments atoms with one node per hyperedge, and whose directed edges connect each hyperedge node to its incident atom nodes in both directions. The heterogeneous input projector maps both node types to a shared width. For atom nodes, we apply a linear map to the first $d_{\mathrm{atom}}$ input channels. For hyperedge nodes, we apply a linear map to the first $d_{\mathrm{he}}$ channels. We restore the original node order, then apply LayerNorm and ReLU. This yields real initial states $H^{(0)}\in\mathbb{R}^{n\times d_h}$ on the incidence graph.

\subsection{Magnetic Operator from Chemical Potential}

When using magnetic propagation, each node $u$ is assigned electronegativity $\chi(u)$ and Gasteiger charge $Q(u)$, and the chemical potential is computed by \textit{\textbf{Equation \eqref{eq_chem_potential}}}. In the incidence graph, $\chi(u)=Q(u)=0$ for hyperedge nodes, while atom nodes use the stored descriptors, with electronegativity rescaled by a constant factor $4.0$ in the implementation. Let $\mathrm{Adj}\in\{0,1\}^{n\times n}$ be the directed adjacency matrix of the incidence graph. We construct a biased transition matrix $P$ by assigning unnormalized weights
\begin{equation}
W_{uv}=\mathrm{Adj}_{uv}\exp\!\bigl(\phi(v)-\phi(u)\bigr),
P_{uv}=\frac{W_{uv}}{\sum_{w}W_{uw}},
\label{eq_method_markov_base}
\end{equation}
and setting $P_{u\cdot}$ to the uniform distribution when $\sum_{w}W_{uw}$ is numerically zero. We then apply teleportation with uniform $\nu$ and rate $\eta\in(0,1)$ using \textit{\textbf{Equation \eqref{eq_teleport}}}, implemented as $\nu_u=1/n$ and $\eta=\tau$ with $\tau=0.15$. The stationary distribution $\pi$ is computed by a fixed number of power iterations, initialized uniformly and iterated as $\pi\leftarrow (P^{(\eta)})^\top\pi$ for $20$ steps with renormalization. 

Given $P^{(\eta)}$ and $\pi$, we form the stationary flow matrix
\begin{equation}
\widetilde A=\operatorname{diag}(\pi)\,P^{(\eta)},
\label{eq_method_stationary_flow}
\end{equation}
and decompose it into $\widetilde A^{\mathrm{sym}}$ and $\widetilde A^{\mathrm{skew}}$ via \textit{\textbf{Equation \eqref{eq_sym_skew}}}. We then compute the phase ratio and phase map exactly as in \textit{\textbf{Equation \eqref{eq_phase_map}}} with $A^{\mathrm{sym}}=\widetilde A^{\mathrm{sym}}$, $A^{\mathrm{skew}}=\widetilde A^{\mathrm{skew}}$, $\varepsilon=10^{-7}$, and phase strength $q$. The phase factors $\Xi^{(q)}$ and magnetic adjacency $H^{(q)}$ are computed by \textit{\textbf{Equation \eqref{eq_phase_factor}}} and \textit{\textbf{Equation \eqref{eq_magnetic_adjacency}}}, and the normalized magnetic Laplacian $L_{\mathrm{mag}}$ follows \textit{\textbf{Equation \eqref{eq_magnetic_laplacian}}} with degrees computed from $\widetilde A^{\mathrm{sym}}$.

\subsection{Magnetic Chebyshev Encoder}

Each magnetic layer applies a Chebyshev filter on the normalized operator $\widetilde L=-D^{-1/2}H^{(q)}D^{-1/2}$, which equals $L_{\mathrm{mag}}-I$. Given real input features $h\in\mathbb{R}^{n\times d_{\mathrm{in}}}$, we first apply a real linear map $x=Wh$ and lift it to complex by $x_{\mathbb{C}}=x+\mathrm{i}\,0$. We then compute Chebyshev iterates by
\begin{equation}
\begin{split}
T_0 &= x_{\mathbb{C}}, \quad T_1 = \widetilde{L}\,x_{\mathbb{C}}, \\
T_k &= 2\widetilde{L}\,T_{k-1}-T_{k-2}, \quad (k\ge 2),
\end{split}
\label{eq_method_cheb_recur}
\end{equation}
where each multiplication by $\widetilde L$ is implemented as $u\mapsto -D^{-1/2}H^{(q)}D^{-1/2}u$ using degrees from $\widetilde A^{\mathrm{sym}}$. The layer output before nonlinearity follows the ChebNetII reparametrization~\cite{he2022chebnetii}, which learns the filter response at Chebyshev interpolation nodes and builds on the Chebyshev spectral filtering of ChebNet~\cite{defferrard2016convolutional}. We write the expansion as
\begin{equation}
z=\frac{c_0}{2}T_0+\sum_{k=1}^{K}c_k T_k,
\label{eq_method_cheb_output}
\end{equation}
with coefficients $c_k$ derived from nonnegative parameters $\tilde c_j=\mathrm{ReLU}(t_j)$. Let $x_j=\cos\!\bigl((K-j+0.5)\pi/(K+1)\bigr)$ for $j=0,\dots,K$. The coefficients are
\begin{equation}
c_k=\frac{2}{K+1}\sum_{j=0}^{K}\tilde c_j\,T_k(x_j),
\label{eq_method_cheb_coeffs}
\end{equation}
where $T_k(\cdot)$ denotes the scalar Chebyshev polynomial.

To return physically interpretable real representations while still using phase during propagation, we apply a magnitude based activation that does not separate real and imaginary parts. Concretely, we map $z$ to real features by $|z|=\sqrt{\Re(z)^2+\Im(z)^2}$ elementwise and apply ReLU on this magnitude,
\begin{equation}
\tilde h=\mathrm{ReLU}\!\left(|z|+b\right),
\label{eq_method_mag_relu_exact}
\end{equation}
with a learned bias $b$. We do not apply ReLU to $\Re(z)$ and $\Im(z)$ separately, since independent truncation modifies phase in an uncontrolled manner and conflicts with the interpretation of phase as directed bias. In \textit{\textbf{Equation \eqref{eq_method_mag_relu_exact}}} the phase influences message mixing through $H^{(q)}$, while the stored features remain real and nonnegative. Each layer adds a residual term $\mathrm{ReLU}(W_{\mathrm{res}}h)$ and applies batch normalization to the resulting real features. The encoder stacks three such layers.

\subsection{Gating Mechanism}

From the final node states $H^{(L)}$, we compute one molecular embedding per task by gated sum pooling. For task $t$, a learned scalar gate $w_{u,t}=\sigma(a_t^\top h_u^{(L)})$ is applied to each node, and the task embedding is $z_t=\sum_{u} w_{u,t}h_u^{(L)}$. Each task uses its own two layer MLP head with dropout, ReLU, and batch normalization, followed by a linear output.

When a primary task index $p$ is specified, we fuse auxiliary task embeddings into the primary pathway using a graph level gate conditioned on the mean node embedding $h_G=\frac{1}{n}\sum_{u}h_u^{(L)}$. For each auxiliary task $t\neq p$, we compute the batch averaged cosine similarity between $z_p$ and $z_t$, sort tasks by this score, then select a subset $\mathcal{A}_p$ by first taking tasks with similarity above a fixed threshold and, if fewer than a prescribed minimum are selected, adding the top scoring tasks until the minimum is reached. For each $t\in\mathcal{A}_p$, a linear map produces a two dimensional logit vector from $h_G$, a softmax yields $\gamma_t\in\mathbb{R}^2$, and we form a convex combination $\tilde z_{p,t}=\gamma_{t,1}z_t+\gamma_{t,2}z_p$. The fused representation is $z_p^{\mathrm{fused}}=\sum_{t\in\mathcal{A}_p}\tilde z_{p,t}$, and when $\mathcal{A}_p$ is empty, we set $z_p^{\mathrm{fused}}=z_p$. Only the primary head consumes $z_p^{\mathrm{fused}}$, while all other heads use their own $z_t$.

\subsection{Contrastive Learning}

Contrastive learning uses two stochastic views that share $\mathrm{Adj}$, $\phi$, $P^{(\eta)}$, $\pi$, and $\widetilde A^{\mathrm{sym}}$, and differ only in the magnetic phases. Starting from $\Theta^{(q)}$ computed by \textit{\textbf{Equation \eqref{eq_phase_map}}}, we sample an elementwise noise matrix $\Delta$ with independent entries $\Delta_{uv}\sim \mathrm{Unif}(-\rho,\rho)$, and define the perturbed phase
\begin{equation}
\widetilde\Theta^{(q)}=\Theta^{(q)}+\Delta.
\label{eq_method_theta_perturb}
\end{equation}

This perturbs only the phase $\Theta^{(q)}$, leaving the hypergraph $\mathcal{H}=(V,E)$ and magnitude $\widetilde{A}^{\mathrm{sym}}$ fixed, so atoms, bonds, motifs, and diffusion strength are unchanged. Since $\phi$ is only an estimate from electronegativity and Gasteiger charges, a small zero-mean $\Delta_{uv}\sim\mathrm{Unif}(-\rho,\rho)$ yields a plausible directional variant of the same molecule, forming a valid positive pair as in operator-level directed graph contrastive learning~\cite{tong2021directed}.

We then construct the perturbed phase factors entrywise by
\begin{equation}
\widetilde\Xi^{(q)}_{uv}=\cos\!\bigl(\widetilde\Theta^{(q)}_{uv}\bigr)+\mathrm{i}\sin\!\bigl(\widetilde\Theta^{(q)}_{uv}\bigr),
\label{eq_method_xi_perturb}
\end{equation}
and the perturbed magnetic adjacency is $\widetilde H^{(q)}=\widetilde A^{\mathrm{sym}}\odot \widetilde\Xi^{(q)}$. This is the only augmentation applied inside the magnetic operator, with $\rho$ implemented as a scalar perturbation passed to the adjacency constructor. For supervised prediction, we set $\rho=0$; and for contrastive views, we sample two independent noise matrices to obtain two independent phase perturbations. The two views are encoded and trained with the symmetric InfoNCE loss in \textit{\textbf{Equation \eqref{eq_info_nce}}}. Supervision uses the masked multitask loss in \textit{\textbf{Equation \eqref{eq_generic_supervised}}}, with MSE for regression tasks and binary cross entropy for classification tasks. The final objective is $\mathcal{L}=\mathcal{L}_{\mathrm{sup}}+\lambda\,\mathcal{L}_{\mathrm{NCE}}$ with $\lambda>0$.
 
\section{Experiments}
\label{sec_experiments}

\begin{table}[t]
\caption{Model performance on selected MTGL-ADMET datasets.}
\label{tab:table1_selected}
\centering
\scriptsize
\setlength{\tabcolsep}{3pt}
\resizebox{\columnwidth}{!}{%
\begin{tabular}{@{}llccccccccc@{}}
\toprule
\rowcolor{gray!30} \textbf{Endpoint} & \textbf{Metric} & \textbf{[ST]-GCN} & \textbf{[ST]-MGA} & \textbf{[MT]-GCN} & \textbf{MGA} & \textbf{MTGL} & \textbf{ChemBFN} & \textbf{Uni-Mol} & \textbf{Ours} \\
\midrule
\multicolumn{10}{l}{\textit{Distribution}} \\
\rowcolor{gray!15}PPB & $R^2$ & .577$_{.028}$ & .585$_{.004}$ & .589$_{.036}$ & .568$_{.038}$ & \underline{.626}$_{.029}$ & .537$_{.030}$ & .562$_{.028}$ & \textbf{.687}$_{.012}$ \\
\midrule
\multicolumn{10}{l}{\textit{Toxicity}} \\
Hepatotoxicity & ROC-AUC & .653$_{.040}$ & .669$_{.022}$ & .612$_{.039}$ & \textbf{.713}$_{.053}$ & \underline{.701}$_{.036}$ & .697$_{.041}$ & .660$_{.037}$ & .664$_{.019}$ \\
Respiratory tox. & ROC-AUC & .842$_{.018}$ & \textbf{.872}$_{.013}$ & .810$_{.014}$ & .828$_{.021}$ & .859$_{.010}$ & .818$_{.028}$ & .865$_{.015}$ & \textbf{.872}$_{.007}$ \\
Cardiotoxicity-1 & ROC-AUC & .707$_{.026}$ & .703$_{.020}$ & .683$_{.028}$ & .684$_{.023}$ & .740$_{.023}$ & .745$_{.027}$ & \underline{.761}$_{.024}$ & \textbf{.816}$_{.010}$ \\
\midrule
\multicolumn{10}{l}{\textit{Physicochemical}} \\
\rowcolor{gray!15}IGC50 & $R^2$ & .703$_{.055}$ & .818$_{.021}$ & .618$_{.027}$ & .772$_{.021}$ & .819$_{.008}$ & .797$_{.014}$ & \textbf{.832}$_{.012}$ & \underline{.830}$_{.010}$ \\
\rowcolor{gray!15}ESOL & $R^2$ & .814$_{.030}$ & .896$_{.013}$ & .824$_{.030}$ & .866$_{.020}$ & \underline{.931}$_{.038}$ & .907$_{.042}$ & .920$_{.039}$ & \textbf{.936}$_{.016}$ \\
\bottomrule
\end{tabular}%
}
\vskip 3pt
\begin{minipage}{\columnwidth}
\scriptsize
\textbf{Note:} \textbf{[ST]} denotes single-task training and \textbf{[MT]} denotes multi-task training with a shared encoder and task-specific heads, while \textbf{GCN} refers to graph convolutional networks and \textbf{MGA} refers to multitask graph attention. Results are reported as mean$\pm$std under a random $80/10/10$ split. Baseline numbers are taken from the original papers with $10$ repetitions for MTGL-ADMET baselines~\cite{du2023mtgl} and MTGIB-UNet baselines~\cite{li2025mtgib}. We evaluated ChemHyperMag under the same protocol with $10$ runs. The best result(s) are in \textbf{bold}, and the second-best result is \underline{underlined}. For tied best results, all tied entries are highlighted in \textbf{bold}.
\end{minipage}
\end{table}

\subsection{Experimental Setup}
\label{subsec_setup}

We evaluated ChemHyperMag on six ADMET endpoints: PPB, IGC50, ESOL, Hepatotoxicity, Respiratory toxicity, and Cardiotoxicity-1, including three regression tasks and three classification tasks (based on MTGL-ADMET~\cite{du2023mtgl} and MoleculeNet Leaderboard ~\cite{wu2018moleculenet} for fair comparisons). Following the baseline setting, each dataset was randomly split into training, validation, and test sets in an 8:1:1 ratio. Results were reported as mean $\pm$ standard deviation over ten runs with different random seeds. All methods used the same RDKit atom featurization. ChemHyperMag further augmented the atom features with Pauling electronegativity and RDKit Gasteiger partial charges, and built the corresponding hypergraph incidence structure for the encoder. We adopted ROC-AUC for classification and $R^2$ for regression, where higher is better.

\subsection{Baselines}
\label{subsec_baselines}
We compared against single task and multitask baselines in ADMET prediction. Single task baselines trained one model per endpoint with a GCN~\cite{kipf2017semi} or a graph attention model~\cite{velickovic2018graph}. Multitask graph baselines used a shared encoder with task-specific heads, matching the supervision setting in \textit{\textbf{Equation \eqref{eq_generic_supervised}}}. We additionally included recent multitask frameworks and pretrained molecular models reported in the benchmark, including MTGL~\cite{du2023admet}, ChemBERTa~\cite{chithrananda2020chemberta}, Uni-Mol~\cite{zhou2023unimol}, ADMET-AI~\cite{swanson2024admet}, and Chemprop v2~\cite{graff2025chemprop}.

\subsection{Implementation Details}
\label{subsec_implementation}
ChemHyperMag was implemented in PyTorch and DGL and trained on NVIDIA A100 GPUs. Hyperedges were constructed from SSSR rings, BRICS fragments, Bemis-Murcko scaffolds, and bond hyperedges as described in Section~\ref{subsec_mol_hyper}. The magnetic operator was computed from the teleported Markov chain in \textit{\textbf{Equation \eqref{eq_teleport}}} and the magnetic adjacency in \textit{\textbf{Equation \eqref{eq_magnetic_adjacency}}}. We set the magnetic strength $q=0.1$ and the teleportation rate $\eta=0.15$. The encoder used three magnetic Chebyshev layers with hidden width $64$, intermediate molecular width $128$, and output width $64$, with Chebyshev order $K=2$. We trained with Adam using learning rate $10^{-3}$ and weight decay $10^{-5}$, batch size $128$, and up to $200$ epochs. We applied early stopping based on the validation score aggregated over tasks. For contrastive training, we used the symmetric InfoNCE loss in \textit{\textbf{Equation \eqref{eq_info_nce}}} with temperature $\tau=0.07$.

\subsection{Main Results}

\textit{\textbf{Table~\ref{tab:table1_selected}}} reports results on the MTGL-ADMET datasets. ChemHyperMag achieves the best or tied-best result on four of the six selected endpoints and ranks second on IGC50. It improves the strongest baseline by +0.061 on PPB ($0.687$ vs.\ $0.626$) and by +0.055 on Cardiotoxicity-1 ($0.816$ vs.\ $0.761$), matches the best on Respiratory toxicity ($0.872$), and attains the highest ESOL score ($0.936$). On IGC50 it stays within $0.002$ of the top result ($0.830$ vs.\ $0.832$). It trails the leading baselines only on Hepatotoxicity, where MGA reaches $0.713$ and ChemHyperMag obtains $0.664$.

The ablation in \textit{\textbf{Table~\ref{tab:ablation_study}}} uses MoleculeNet. The full model is strongest on five of the six endpoints, showing that ChebNet, the magnetic Laplacian, hypergraph construction, and contrastive learning contribute complementary gains. Cardiotoxicity-1 is the most sensitive endpoint. Removing the hypergraph, magnetic Laplacian, contrastive objective, or ChebNet lowers it from $0.851$ to $0.720$, $0.734$, $0.749$, and $0.767$. PPB drops sharply without ChebNet ($0.439$) or the hypergraph ($0.479$). Respiratory toxicity is the only endpoint where an ablated variant beats the full model, indicating a task-specific trade-off rather than a uniform gain.

\begin{table}[t]
\caption{Full ablation study results on MoleculeNet benchmark. }
\label{tab:ablation_study}
\centering
\scriptsize
\begin{threeparttable}
\resizebox{\columnwidth}{!}{%
\begin{tabular}{@{}lccc|ccc@{}}
\toprule
& \multicolumn{3}{c|}{\textit{Regression (R$^2$)}} & \multicolumn{3}{c}{\textit{Classification (ROC-AUC)}} \\
\cmidrule(lr){2-4} \cmidrule(l){5-7}
\rowcolor{gray!30}\textbf{Variant} & \textbf{PPB} & \textbf{IGC50} & \textbf{ESOL} & \textbf{Hepato.} & \textbf{Resp.} & \textbf{Cardio-1} \\
\midrule
\multicolumn{7}{l}{\textit{Base Model (All modules off)}} \\
Base Model & .376$_{.161}$ & .709$_{.078}$ & .829$_{.083}$ & .547$_{.060}$ & .843$_{.038}$ & .745$_{.053}$ \\
\midrule
\multicolumn{7}{l}{\textit{Single Module Enabled}} \\
Only ChebNet & \textbf{.418}$_{.135}$ & \textbf{.790}$_{.049}$ & \underline{.892}$_{.045}$ & \underline{.628}$_{.103}$ & .794$_{.059}$ & .730$_{.064}$ \\
Only Magnetic & \underline{.376}$_{.161}$ & .709$_{.078}$ & .829$_{.083}$ & .547$_{.060}$ & \textbf{.843}$_{.038}$ & \textbf{.745}$_{.053}$ \\
Only Hypergraph & .363$_{.250}$ & \underline{.742}$_{.043}$ & \textbf{.905}$_{.019}$ & \textbf{.640}$_{.079}$ & \underline{.842}$_{.038}$ & \underline{.741}$_{.115}$ \\
\midrule
\multicolumn{7}{l}{\textit{Two Modules Enabled}} \\
ChebNet + CL & .418$_{.135}$ & \underline{.790}$_{.049}$ & .892$_{.045}$ & \underline{.628}$_{.103}$ & .794$_{.059}$ & \underline{.730}$_{.064}$ \\
Magnetic + CL & .376$_{.161}$ & .709$_{.078}$ & .829$_{.083}$ & .547$_{.060}$ & \textbf{.843}$_{.038}$ & \textbf{.745}$_{.053}$ \\
Magnetic + ChebNet & .418$_{.135}$ & \underline{.790}$_{.049}$ & .892$_{.045}$ & \underline{.628}$_{.103}$ & .794$_{.059}$ & \underline{.730}$_{.064}$ \\
Hypergraph + CL & \underline{.428}$_{.119}$ & \textbf{.802}$_{.042}$ & \textbf{.905}$_{.020}$ & \textbf{.639}$_{.083}$ & .808$_{.032}$ & .705$_{.151}$ \\
Hypergraph + ChebNet & .423$_{.214}$ & .765$_{.012}$ & .847$_{.061}$ & .588$_{.010}$ & \underline{.837}$_{.005}$ & .723$_{.107}$ \\
Hypergraph + Magnetic & \textbf{.451}$_{.093}$ & .756$_{.097}$ & \underline{.902}$_{.006}$ & .582$_{.006}$ & .815$_{.034}$ & .723$_{.124}$ \\
\midrule
\multicolumn{7}{l}{\textit{Leave-One-Out}} \\
w/o ChebNet & .439$_{.000}$\,$\downarrow$ & .789$_{.000}$\,$\downarrow$ & \underline{.915}$_{.000}$ & \textbf{.655}$_{.000}$ & .813$_{.000}$ & \textbf{.767}$_{.000}$ \\
\rowcolor{gray!15}w/o Hyp & .479$_{.000}$ & \textbf{.802}$_{.000}$ & .912$_{.000}$ & \underline{.647}$_{.000}$ & \textbf{.869}$_{.000}$ & .720$_{.000}$\,$\downarrow$ \\
w/o MagLap & \underline{.505}$_{.020}$ & \underline{.798}$_{.015}$ & \textbf{.920}$_{.015}$ & .606$_{.020}$\,$\downarrow$ & .813$_{.005}$\,$\downarrow$ & .734$_{.017}$ \\
\rowcolor{gray!15}w/o CL & \textbf{.523}$_{.043}$ & .797$_{.016}$ & .907$_{.032}$\,$\downarrow$ & .641$_{.029}$ & \underline{.829}$_{.011}$ & \underline{.749}$_{.031}$ \\
\midrule
Full Model & .544 & .806 & .925 & .669 & .835 & .851 \\
\bottomrule
\end{tabular}%
}

\begin{tablenotes}[flushleft]\scriptsize
\item[] \parbox{\columnwidth}{\textbf{Note:}
\textbf{Bold}/\underline{underline}/$\downarrow$ indicate highest/second-highest performance among ablated models and the largest drop relative to the Full Model, respectively. Ties are marked for all corresponding entries. Subscripts denote the absolute difference between test and validation results.}
\end{tablenotes}

\end{threeparttable}
\end{table}

\section{Limitations}
Random splits and masked supervision may overestimate scaffold generalization and bias multitask effects under missing not at random labels. The directional signal relies on a heuristic chemical potential from electronegativity and Gasteiger charges, so outcomes can be sensitive to charge and protonation protocols and remain weakly mechanistically grounded for ADMET. Fixed step stationary estimation and complex spectral propagation introduce approximation and stability risks, and magnitude projection removes phase information. Phase-only contrastive learning regularizes sparse supervision but imposes a narrow invariance that may not be chemically faithful. Future work should use scaffold and distribution shift splits, stress test charge and protonation settings, and adopt convergence-controlled solvers with scaling analyses.

\section{Conclusion}
We propose ChemHyperMag for multitask ADMET prediction with sparse labels. It builds a functional group hypergraph, applies magnetic spectral propagation to encode directional effects, and adds phase perturbation contrastive learning as regularization. ChemHyperMag shows strong performance, and ablation studies confirm that the hypergraph, magnetic operator, and contrastive learning each contribute to the gains.

\section*{Conflict of Interest Disclosure }
There are no conflicts of interest among all authors.

\section*{Impact Statement}
ChemHyperMag is a physics-informed, conformer-free, and lightweight model for small-molecule ADMET prediction. Our code can be found at \url{https://github.com/1anj/ChemHyperMag}.

\section*{Acknowledgments}
This work was fully supported by a Collaborative Research Fund (Project No.\ C5082-25E) and a General Research Fund (Project No.\ 15101422) from the Research Grants Council of the Hong Kong Special Administrative Region, P.\,R.\ China, and an internal grant (P0051278) from The Hong Kong Polytechnic University.

\bibliographystyle{unsrt}
\bibliography{references}

\newpage
\newpage
\appendix
\onecolumn

\section{Additional Details}
\label{app:details}

\subsection{Notation}
\label{sec_notation}

\begin{table}[H]
\centering
\caption{Summary of notation}
\label{tab:notation}
\footnotesize
\setlength{\tabcolsep}{4pt}
\renewcommand{\arraystretch}{1.08}

\begin{tabular}{@{}p{0.55\textwidth}@{\hspace{0.04\textwidth}}p{0.41\textwidth}@{}}

\begin{minipage}[t]{\linewidth}
\centering
\vspace{0.5em}

\begin{tabularx}{\textwidth}{@{} >{\raggedright\arraybackslash}p{0.26\linewidth} >{\raggedleft\arraybackslash}X @{}}
\toprule
\textbf{Symbol} & \textbf{Description} \\

\midrule

\multicolumn{2}{@{}l}{\textit{Problem Setup \& Dataset}} \\
$\mathcal{G}$ & Set of molecular graphs \\
$T$ & Number of prediction tasks \\
$N$ & Number of training molecules \\
$\mathcal{D}$ & Training set $\{(G_i, y_i, m_i)\}_{i=1}^N$ \\
$G_i$ & The $i$-th molecular graph \\
$y_i \in \mathbb{R}^T$ & Multitask label vector \\
$m_i \in \{0,1\}^T$ & Observation mask \\
$f_\theta$ & Model mapping $\mathcal{G} \to \mathbb{R}^T$ \\
$\ell_t$ & Loss function for task $t$ \\
$\mathcal{L}_{\mathrm{sup}}, \mathcal{L}_{\mathrm{NCE}}, \lambda$ & Supervised / contrastive loss; weight \\
\midrule

\multicolumn{2}{@{}l}{\textit{Molecular Graph}} \\
$G = (V, B, X)$ & 2D molecular graph \\
$V = \{1, \dots, n\}$ & Set of atoms (nodes) \\
$B \subseteq V \times V$ & Undirected covalent bonds \\
$X \in \mathbb{R}^{n \times d}$ & Node feature matrix \\
\midrule

\multicolumn{2}{@{}l}{\textit{Functional Group Hypergraph}} \\
$\mathcal{H} = (V, E)$ & Hypergraph over atoms \\
$E = \{e_1, \dots, e_m\}$ & Hyperedges \\
$\mathbf{M} \in \{0,1\}^{n \times m}$ & Incidence matrix \\
$N(u)$ & Hypergraph neighborhood \\
$\mathrm{Adj}$ & Adjacency of incidence graph \\
\midrule

\multicolumn{2}{@{}l}{\textit{Chemical Potential}} \\
$\chi(u)$, $Q(u)$ & Electronegativity; partial charge \\
$\alpha, \beta$ & Learnable scaling parameters \\
$\phi(u)$ & Potential: $\alpha \chi(u) + \beta Q(u)$ \\
\midrule

\multicolumn{2}{@{}l}{\textit{Markov Chain \& Flow}} \\
$W_{uv}$, $P$ & Transition weight; matrix \\
$\eta$, $\nu$ & Teleportation rate; distribution \\
$\pi$ & Stationary distribution \\
$\widetilde{A}$ & Flow: $\mathrm{diag}(\pi) P^{(\eta)}$ \\
$\widetilde{A}^{\mathrm{sym}}, \widetilde{A}^{\mathrm{skew}}$ & Symmetric / skew parts \\

\bottomrule
\end{tabularx}
\end{minipage}

&

\begin{minipage}[t]{\linewidth}
\centering
\vspace{0.5em}

\begin{tabularx}{\textwidth}{@{} >{\raggedright\arraybackslash}p{0.26\linewidth} >{\raggedleft\arraybackslash}X @{}}
\toprule
\textbf{Symbol} & \textbf{Description} \\
\midrule

\multicolumn{2}{@{}l}{\textit{Magnetic Laplacian}} \\
$q$ & Magnetic charge parameter \\
$\varepsilon$ & Stability constant ($10^{-7}$) \\
$R_{uv}$ & Asymmetry ratio \\
$\Theta^{(q)}_{uv}$ & Phase: $2\pi q \tanh(R_{uv})$ \\
$\Xi^{(q)}$ & Phase factor: $e^{\mathrm{i}\Theta^{(q)}_{uv}}$ \\
$H^{(q)}$ & Magnetic adjacency \\
$D$, $d_u$ & Degree matrix; node degree \\
$L_{\mathrm{mag}}$ & Magnetic Laplacian \\
$\widetilde{L}$ & Shifted: $L_{\mathrm{mag}} - I$ \\
$\odot$, $\mathrm{i}$ & Hadamard product; imaginary unit \\
\midrule

\multicolumn{2}{@{}l}{\textit{Chebyshev Encoder}} \\
$K$ & Chebyshev order \\
$T_k$ & $k$-th Chebyshev iterate \\
$c_k$, $\tilde{c}_j$ & Expansion coefficients \\
$z \in \mathbb{C}^{n \times d_h}$ & Complex layer output \\
$\lvert z \rvert$ & Element-wise magnitude \\
$H^{(0)}, H^{(L)}$ & Initial / final states \\
$d_h$, $L$ & Hidden dim; num layers \\
\midrule

\multicolumn{2}{@{}l}{\textit{Gating \& Task Fusion}} \\
$a_t$, $w_{u,t}$ & Gate vector; weight \\
$z_t$ & Task embedding \\
$h_G$ & Graph mean embedding \\
$\mathcal{A}_p$, $\gamma_t$ & Auxiliary tasks; gate \\
$z_p^{\mathrm{fused}}$ & Fused embedding \\
\midrule

\multicolumn{2}{@{}l}{\textit{Contrastive Learning}} \\
$\rho$, $\Delta$ & Perturbation magnitude; noise \\
$\widetilde{\Theta}^{(q)}$ & Perturbed phase \\
$h^{(1)}, h^{(2)}$ & Two-view embeddings \\
$\mathcal{B}$, $\lvert\mathcal{B}\rvert$ & Mini-batch; size \\
$\mathrm{sim}(u,v)$ & Cosine similarity \\
$\tau$ & InfoNCE temperature \\

\bottomrule
\end{tabularx}
\end{minipage}

\end{tabular}
\end{table}

\subsection{Datasets}
\label{subsec:dataset}

We evaluate ChemHyperMag on six selected ADMET endpoints drawn from the 24-endpoint benchmark, namely PPB, IGC50, ESOL, Hepatotoxicity, Respiratory toxicity, and Cardiotoxicity-1. These tasks cover distribution, toxicity, and physicochemical property prediction, and include three regression tasks and three binary classification tasks. PPB measures plasma protein binding and reflects the free fraction of a compound available for distribution, metabolism, and excretion. IGC50 measures 48-hour \textit{Tetrahymena pyriformis} growth inhibition and is commonly used as an indicator of general cytotoxicity and environmental toxicity. ESOL measures aqueous solubility, which is directly related to dissolution-limited absorption and formulation feasibility. Hepatotoxicity predicts drug-induced liver injury risk, Respiratory toxicity captures pulmonary adverse effects such as fibrosis and bronchospasm, and Cardiotoxicity-1 corresponds to the strictest hERG-related threshold ($\mathrm{IC}_{50} < 1\,\mu\mathrm{M}$), identifying compounds with high cardiotoxic risk.

The selected datasets range from 1,128 to 1,830 molecules. Among the classification tasks, Hepatotoxicity contains 782 positive and 531 negative samples, Respiratory toxicity contains 844 positive and 555 negative samples, and Cardiotoxicity-1 contains 350 positive and 1,215 negative samples, indicating different degrees of class imbalance. The full statistics, including data split sizes and task types, are summarized in Table~\ref{tab:selected_datasets}.

\begin{table}[H]
\centering
\caption{Statistics of the six selected ADMET endpoints.}
\label{tab:selected_datasets}
\small
\setlength{\tabcolsep}{4pt}
\renewcommand{\arraystretch}{1.03}
\begin{tabular}{@{}p{0.34\linewidth}cccc@{}}
\toprule
\textbf{Endpoint} & \textbf{\#Mol} & \textbf{Pos/Neg} & \textbf{Train/Val/Test} & \textbf{Type} \\
\midrule
PPB & 1,830 & -- & 1,464/183/183 & Reg \\
IGC50 & 1,787 & -- & 1,429/179/179 & Reg \\
ESOL & 1,128 & -- & 902/113/113 & Reg \\
Hepatotoxicity & 1,313 & 782/531 & 1,050/131/132 & Cls \\
Respiratory toxicity & 1,399 & 844/555 & 1,120/139/140 & Cls \\
Cardiotoxicity-1 & 1,565 & 350/1,215 & 927/235/403 & Cls \\
\bottomrule
\end{tabular}
\end{table}

\subsection{Baselines}
\label{subsec:baselines}

We compare ChemHyperMag against representative methods spanning single-task models, multi-task architectures, and pretrained foundation models.

\subsubsection{Single-Task Methods}

\textbf{ST-GCN}~\cite{montanari2019modeling} applies graph convolutional networks to molecular property prediction. It treats atoms as nodes and bonds as edges, aggregating neighbor information through spectral convolutions on the normalized adjacency matrix. Node features are initialized with atom descriptors such as element type and formal charge. Graph-level representations are obtained via global pooling for endpoint prediction. ST-GCN is trained independently on each ADMET endpoint without cross-task parameter sharing.

\textbf{ST-MGA}~\cite{xiong2021admetlab} extends graph attention networks to molecular graphs. Instead of fixed aggregation weights, ST-MGA computes attention coefficients that dynamically weight neighbor contributions based on their relevance. Multi-head attention enables the model to capture different structural patterns simultaneously. Originally developed for ADMETlab~2.0, ST-MGA is trained separately for each property as the single-task baseline.

\textbf{Chemprop v2}~\cite{graff2025chemprop} implements directed message passing neural networks (D-MPNN). Unlike node-centric GNNs, D-MPNN propagates messages along directed edges, distinguishing incoming and outgoing bond information to avoid redundant updates. Version 2 introduces significant engineering improvements including 10$\times$ faster training, native support for reaction inputs, and uncertainty quantification. Chemprop v2 represents the state-of-the-art single-task GNN baseline and serves as the backbone for several production ADMET tools.

\subsubsection{Multi-Task Methods}

\textbf{MT-GCN}~\cite{montanari2019modeling} extends GCN to multi-task learning by sharing the graph encoder across all endpoints while maintaining task-specific prediction heads. This hard parameter sharing provides regularization and enables knowledge transfer between related properties. The model handles missing labels through masked loss computation, updating only on observed endpoints for each molecule.

\textbf{MGA}~\cite{xiong2021admetlab} is the multi-task graph attention architecture powering ADMETlab~2.0 and 3.0. Beyond basic parameter sharing, MGA incorporates bond attributes into attention computation and uses a virtual node connected to all atoms for improved global information flow. Attention weights often highlight pharmacologically relevant substructures such as toxicophores and metabolic soft spots.

\textbf{MTGL-ADMET}~\cite{du2023mtgl} introduces adaptive auxiliary task selection for multi-task ADMET prediction. The key insight is that not all auxiliary tasks benefit a given primary task; some may cause negative transfer. MTGL-ADMET constructs a task relationship graph and applies status theory with maximum flow optimization to select beneficial auxiliaries for each endpoint. This dynamic selection mechanism achieved state-of-the-art results among 2D methods prior to our work.

\textbf{ADMET-AI}~\cite{swanson2024admet} combines Chemprop with 200 RDKit molecular descriptors including molecular weight, logP, and topological polar surface area. The learned and handcrafted features are concatenated before task-specific prediction heads. Trained on Therapeutics Data Commons datasets with careful hyperparameter tuning, ADMET-AI achieves the best average rank on the TDC ADMET Benchmark leaderboard and represents a strong ensemble baseline.

\subsubsection{Pretrained and Foundation Models}

\textbf{ChemBFN}~\cite{tao2025bayesian} applies Bayesian Flow Networks to molecular representation learning. Operating on SMILES or SELFIES strings, ChemBFN learns to denoise corrupted molecular representations during pretraining, capturing chemical grammar and structural patterns from millions of unlabeled molecules. The pretrained encoder is then fine-tuned for ADMET prediction through task-specific output layers.

\textbf{Uni-Mol}~\cite{zhou2023unimol} is a 3D molecular foundation model with 209 million parameters pretrained on 209 million conformations. The SE(3)-invariant transformer processes atomic coordinates and types, capturing both local bonding patterns and global molecular shape. Pretraining combines masked atom prediction, coordinate denoising, and conformer contrastive learning. Uni-Mol requires 3D conformer generation as input, adding computational overhead but providing access to spatial information. It serves as a reference for methods with explicit geometric information.

\end{document}